\documentclass[svgnames]{scrartcl}

\usepackage{url}            
\usepackage{booktabs}       
\usepackage{nicefrac}       
\usepackage{microtype}      
\usepackage{graphicx}
\usepackage{multirow}

\usepackage{natbib}

\usepackage{amsmath}
\usepackage{amsfonts}
\usepackage{amssymb}
\usepackage{amsthm}

\usepackage{xcolor}

\usepackage{iftex}
\ifPDFTeX
   \usepackage[utf8]{inputenc}
   \usepackage{lmodern}
   \usepackage[T1]{fontenc}

\else
   \ifXeTeX
     \usepackage{fontspec}
   \else 
     \usepackage{luatextra}
   \fi
   \defaultfontfeatures{Ligatures=TeX}
\fi

\title{Dictionary Learning for Robotic Grasp Recognition and Detection}

\author{
  Ludovic Trottier \qquad  Philippe Gigu\`ere  \qquad  Brahim Chaib-draa \\
  Department of Computer Science and Software Engineering \\
  Laval University, Qu\'ebec, Canada \\
  \texttt{ludovic.trottier.1@ulaval.ca} \\
  \texttt{philippe.giguere@ift.ulaval.ca } \\
  \texttt{brahim.chaib-draa@ift.ulaval.ca } \\
}

\date{}

\begin{document}



\maketitle

\begin{abstract}

The ability to grasp ordinary and potentially never-seen objects is an important feature in both domestic and industrial robotics. For a system to accomplish this, it must autonomously identify grasping locations by using information from various sensors, such as Microsoft Kinect 3D camera. Despite numerous progress, significant work still remains to be done in this field. To this effect, we propose a dictionary learning and sparse representation (DLSR) framework for representing RGBD images from 3D sensors in the context of determining such good grasping locations. In contrast to previously proposed approaches that relied on sophisticated regularization or very large datasets, the derived perception system has a fast training phase and can work with small datasets. It is also theoretically founded for dealing with masked-out entries, which are common with 3D sensors. We contribute by presenting a comparative study of several DLSR approach combinations for recognizing and detecting grasp candidates on the standard Cornell dataset. Importantly, experimental results show a performance improvement of 1.69\% in detection and 3.16\% in recognition over current state-of-the-art convolutional neural network (CNN). Even though nowadays most popular vision-based approach is CNN, this suggests that DLSR is also a viable alternative with interesting advantages that CNN has not.

\end{abstract}

\section{Introduction}
In robotics, automating the grasping of ordinary objects is an important open problem~\citep{redmonRealtimeCnnGrasp}. Much progress has been made both on the hardware side (the gripper itself) and on the perception side. The development of compliant or under-actuated mechanical grippers --- often referred to as ``mechanical intelligence'' --- which passively adapt their shape to the grasped object has greatly simplified the problem~\citep{laliberte2002underactuation}. However, determining good grasping locations still requires an efficient perception system. 

The advent of Microsoft Kinect inexpensive 3D camera opened the door to rapid deployment of new and robust approaches in identifying such locations. Its market accessibility and ease-of-use provided a straightforward solution for incorporating depth and RGB information (called RGBD images) into deployed systems of various industrial settings.

\begin{figure}[t]
\centering
\includegraphics[width=0.7\linewidth]{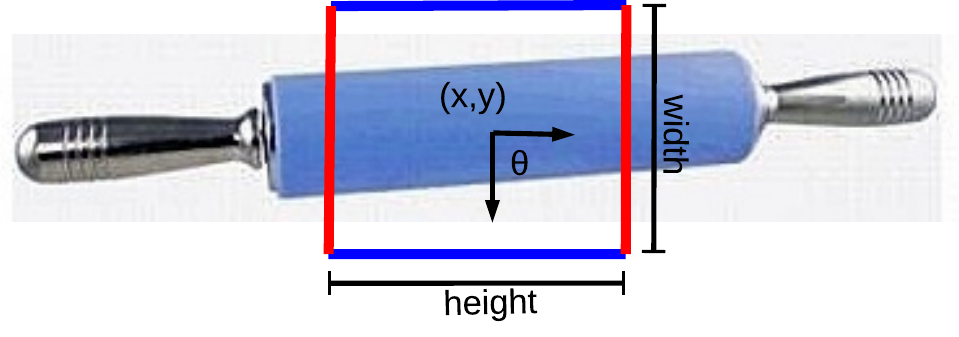}
\caption{The grasp rectangle is a five-dimensional grasp representation. The 2D rectangle is fully determined by its center coordinates ($x,y$), width, height and its angle $\theta$ from the x-axis. The blue edges indicate the gripper plate location and the red edges show the gripper opening, prior to grasping.}
\label{fig:grasping_rectangle}
\end{figure}

In this paper, we look at identifying grasping locations for two plates parallel grippers, by employing such RGBD images. As representation of a grasping location, we use Jiang \textit{et al.} 5-dimensional \textit{grasp rectangle}~\citep{jiang2011efficient} from which the 7-dimensional gripper configuration can be easily computed. The 2D oriented rectangle, shown in Fig.~\ref{fig:grasping_rectangle}, indicates the gripper's location, orientation and physical limitations:
\begin{align*}
R &= \{x,y,\theta, w,h\} \, .
\end{align*}
Using the grasp rectangle representation makes grasp recognition analogous to object recognition (bounding box approaches), and soon for detection. For grasp recognition, the goal is to determine whether a grasp rectangle $R$ is a good or bad candidate. For grasp detection, the goal is to predict the configuration of the best rectangle $R^*$. In this particular setting, identifying grasping locations can then be seen as a vision problem. This is particularly advantageous because several break-through works on similar vision-oriented problems have been proposed in past decades, and can thus be exploited for detecting grasping locations.

Although compelling due to its small cost and ease-of-use, the Microsoft Kinect (and most structured-light devices) has some drawbacks. One is the presence of two types of noise in depth information: the axial and lateral noise model of the object distance to the camera, and the mask noise model of missing 3D information. While vision-based approaches can reasonably deal with the former, the latter can be particularly cumbersome. Objects with shiny surfaces often cause structured-light 3D cameras to fail which results in absence of information. To cope with this phenomenon,~\citet{lenz2015deep} had to develop different mask-based regularization terms to solve their multi-layer neural network convergence problems caused by using zeros as masked-out entry values. With our approach, we seek to create a theoretically founded grasp localization model which can address the Kinect mask noise inherently, without resorting to a custom regularization.

The second drawback of Microsoft Kinect is the lack of available large scale RGBD grasp datasets, which makes training high-dimensional models cumbersome. As an example,~\citet{redmonRealtimeCnnGrasp} previously applied a convolutional neural network (CNN) for detecting grasp candidates that contained more than a million parameters. Due to the relative small amount of RGBD images in their grasp dataset (the Cornell dataset), they had to pre-train the CNN for several days on ImageNet (a RGB image dataset) and fine-tune for several hours on the Cornell one. In an industrial context where datasets are small and new objects are regularly added, a fast and robust training phase is essential. In particular, an efficient strategy to make objects easier to grasp is to add more images from different viewpoints and retrain, hence the importance of fast training phase.

To satisfy the aforementioned requirements, we looked at employing dictionary learning and sparse representations (DLSR). Sparse modeling of data is a biologically-inspired and theoretically founded approach in which observations are represented as linear combinations of few atoms from a dictionary. As previously shown in the context of object recognition and image restoration, DLSR is well-suited to deal with masked-out entries, has a significantly faster training phase than CNN and can work with small datasets~\citep{wright2010sparse}. A standard DLSR method is divided into a \textit{dictionary learning} phase, where a dictionary is trained to capture the latent structure of the data, and a \textit{feature coding} phase, where the dictionary is used to transform raw observations into features. Representing observations by learning how to extract features from them makes DLSR particularly interesting in the actual context, as it steers clear of relying on expert knowledge brought by hand-designed feature engineering.

\begin{figure}[t]
\centering
\includegraphics[width=0.8\linewidth]{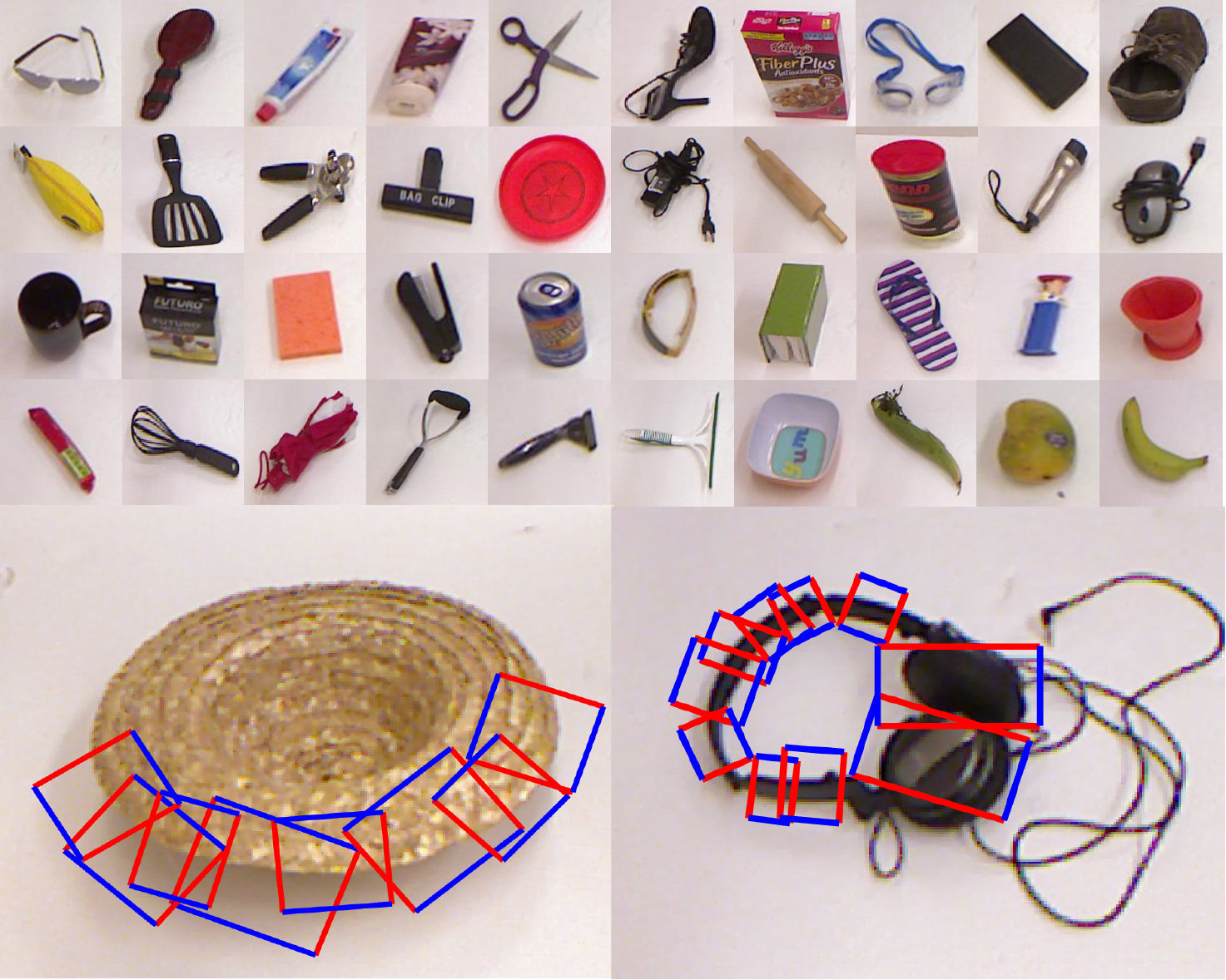}
\caption{The Cornell Grasping Dataset contains images of a wide variety of everyday objects. This database defines two tasks: grasp recognition and grasp detection. For grasp recognition, the goal is to determine whether a grasp rectangle $R$ is a good or a bad candidate. For grasp detection, the goal is to predict the configuration of the best grasp rectangle $R^*$.}
\label{fig:cornell_dataset}
\end{figure}

Our contribution with this paper is twofold. First, we propose a DLSR-based framework for learning and extracting useful information from RGBD images that is rapidly trainable, can work with a small dataset and inherently deals with masked-out entries. The goal is to demonstrate the applicability of such an approach for grasp recognition and detection, and to compare it with other ones in the literature on the standard Cornell task (shown in Fig.~\ref{fig:cornell_dataset}). Second, we present an empirical evaluations of several dictionary learning and feature coding approach combinations. Since DLSR has been around for some years, more than one variant exists for dictionary training and feature extraction. The large quantity of available methods makes choosing one particular combination troublesome, as few indications can guide our choice. By understanding the relationship between dictionary learning and feature coding, our goal is to ascertain which combinations are best suited for the task at hand by comparing performance, speed of training and parallelizability (either on CPU or GPU).

The rest of this paper is divided as follows. We make an overview of related works in section~\ref{sec:related_work}. All dictionary learning approaches and encoders are detailed in section~\ref{sec:learning_framework}, along with explanations concerning data preprocessing and the overall feature extraction process. We elaborate on the experimental framework in section~\ref{sec:experimental_framework} and report the results in section~\ref{sec:experimental_results}. Finally, we discuss the pros and cons of the approaches in section~\ref{sec:discussion} and conclude in section~\ref{sec:conclusion}.

\section{Related Work}
\label{sec:related_work}
One of the fundamental concept in grasping is its representation, which has undergone significant evolution over the years. For example,~\citet{saxena2006robotic} proposed in 2006 a 2D grasping point representation, while more recently,~\citet{le2010learning} proposed a pair of points. However, these representations did not faithfully represent the 7-dimensional gripper configuration and its inherent mechanical constraints, which led to the formulation of the grasp rectangle of~\citet{jiang2011efficient} in 2011.

For identification, several previous approaches used 3D simulations to learn good grasping regions~\citep{goldfeder2007grasp, miller2004graspit, detry2013learning, pelossof2004svm}. A strong limitation for them is the need to know all 3D physical models \emph{a priori}, which severely reduces the applicability for general purpose robots. Better approaches for performing grasp identification without building complex models prior to the execution would certainly be more adequate. 

Other works have shown the importance of depth information~\citep{lai2011large, blum2012learned} and image processing for representing the image inputs~\citep{maitin2010cloth, saxena2008robotic}. They often rely on hand-designed features~\citep{rusu2010fast}, making them possibly brittle or hard to tune. While there has been some work on applying neural network for learning RGBD image features~\citep{socher2012convolutional, gupta2014learning}, robotic grasping using neural networks research is still in its infancy~\citep{lenz2015deep, redmonRealtimeCnnGrasp}.

The literature on Unsupervised Feature Learning (UFL) is vast. DLSR-based approaches like~\citep{aharon2006img, mairal2009online, yang2009linear} achieved impressive results on vision-related tasks such as object recognition~\citep{bo2013unsupervised}, face recognition~\citep{zhang2010discriminative}, scene analysis~\citep{lazebnik2006beyond} and image restoration~\citep{elad2006image}. Conversely, not as much work has been done to evaluate the performance of dictionary learning approaches on RGBD images~\citep{bo2013unsupervised}. By applying such a paradigm for identifying grasping locations, one contribution of this paper is to present a comparative study of several DLSR approach combinations which is currently lacking in grasping literature.

\section{Learning Framework}
\label{sec:learning_framework}

In this section, we detail the dictionary learning approaches that are used for unsupervised feature learning. We also elaborate on the data preprocessing, the overall feature extraction process, the classifier for grasp recognition and the regressor for grasp detection.

\subsection{Data Preprocessing}

\begin{figure}[t]
\centering
\includegraphics[width=\linewidth]{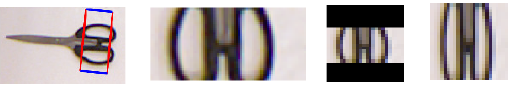}
\caption{Left: a pair of scissors with a candidate grasp rectangle image. Center-Left: image taken from the rectangle rotated to match the global image orientation. Center-Right: rescaled image with preserved aspect ratio. The black regions indicate masked-out padding. Right: rescaled image without preserved aspect ratio. Preserving aspect ratio and using padding allows the object parts to correctly appear graspable.}
\label{fig:imgAspectRatioAll}
\end{figure}

\begin{figure}[t]
    \centering
    \begin{minipage}{.3\linewidth}
        \centering
        \includegraphics[width=\linewidth]{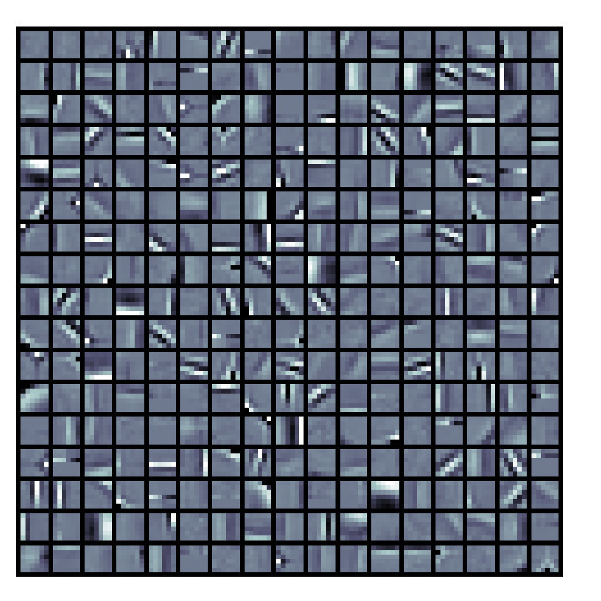}
        {\scriptsize $K$}
    \end{minipage}%
    \begin{minipage}{0.3\linewidth}
        \centering
        \includegraphics[width=\linewidth]{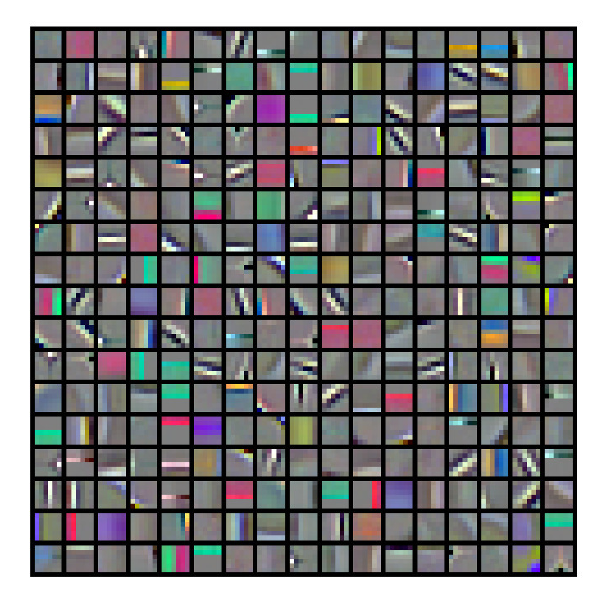}
        {\scriptsize $RGB$}
    \end{minipage}%
    
    \begin{minipage}{0.3\linewidth}
        \centering
        \includegraphics[width=\linewidth]{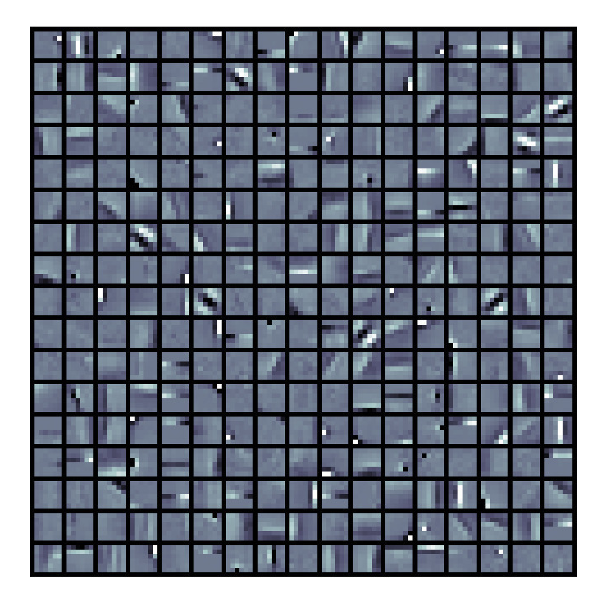}
        {\scriptsize $D$}
    \end{minipage}%
    \begin{minipage}{0.3\linewidth}
        \centering
        \includegraphics[width=\linewidth]{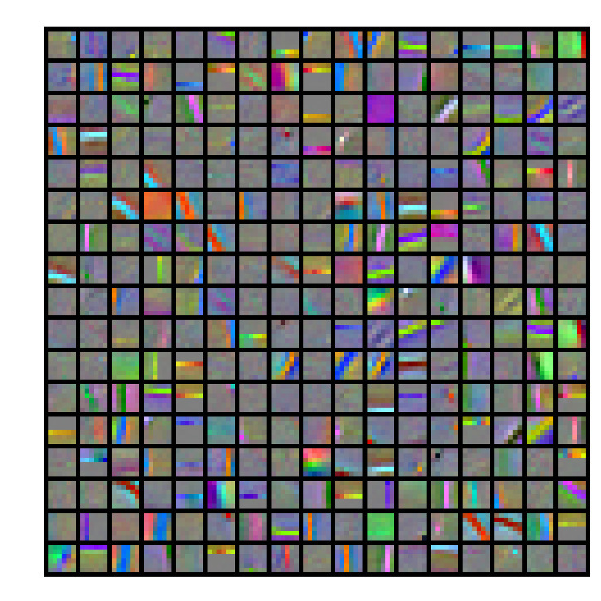}
        \vspace{-3pt}
        {\scriptsize  $N_x N_y N_z$}
    \end{minipage}%
    \caption{A dictionary $D$ of 300 atoms (each square is an atom $D^{(j)}$) learned using the Cornell dataset shown in four distinct parts: K (gray), RGB, D (depth) and $N_x N_y N_z$ (depth normals). Most squares (atoms) show localized and oriented Gabor-like filters.}
    \label{fig:dictionary}
\end{figure}

We first compute the gray channel (K) and estimate the depth normal coordinates $N_x$, $N_y$ and $N_z$ for each RGBD Kinect image, as done in~\citet{bo2013unsupervised}. Each image now contains eight channels. Then, we preprocess each grasp rectangle image by rotating it to match the global image orientation and by rescaling it to a $24 \times 24 \times 8$ size with aspect-ratio preserved.  Fig.~\ref{fig:imgAspectRatioAll} shows an example of a grasp rectangle image that is preprocessed in such a way.

In order to learn a set of features, we first collect a batch (100,000 in our experiments) of small $6 \times 6 \times 8$ patches, extracted at random from the $24 \times 24 \times 8$ images, that are then channel-wise standardized and ZCA whitened~\citep{hyvarinen2004independent}. Given a set of these patch vectors $x^{(i)} \in \mathbb{R}^n$, we apply a dictionary learning approach to learn a dictionary $D \in \mathbb{R}^{n \times d}$, where each column $D^{(j)}$ is one atom that represents the latent structure of the patches. Fig.~\ref{fig:dictionary} shows an example of a dictionary learned with Cornell database, the same we used in our tests. Most squares (atoms) show localized and oriented Gabor-like filters that are known to be relevant features for representing raw images~\citep{marvcelja1980mathematical}.

\subsection{Dictionary Learning}
\label{ssec:dl}

We now elaborate on the dictionary learning algorithms that we chose for learning a dictionary $D$. Specifically, we tested the following approaches:

\subsubsection{Sparse Coding (SC)} 

We train the dictionary by optimizing a $\ell_1$-regularized sparse coding formulation:
\begin{align}
\label{eq:dl_sc}
& \min_{D, w^{(i)}} \sum_{i} \| D w^{(i)} - x^{(i)} \|_2^2 + \lambda \| w^{(i)} \|_1 \\
& \text{subject to } \|D^{(j)}\|_2 = 1, \, \forall j  \nonumber \, ,
\end{align}
using the online dictionary learning (ODL) algorithm of~\citet{mairal2009online}. ODL minimizes~\eqref{eq:dl_sc} alternatively over the sparse weights $w^{(i)}$ and the dictionary $D$, making it a fast and scalable approach for large datasets. We used Least Angle Regression (LARS)~\citep{efron2004least} to solve for the sparse weights. In our experiments, we cross-validated the sparsity parameter $\lambda$ with $\lambda \in \{0.5, 1.0, 1.5, 2.0\}$.

\subsubsection{Orthogonal Matching Pursuit (OMP)}

In this case, the $\ell_1$ penalty is replaced by a $\ell_0$ one, and optimization follows a formulation similar to SC:
\begin{align}
\label{eq:dl_omp}
& \min_{D, w^{(i)}} \sum_{i} \| D w^{(i)} - x^{(i)} \|_2^2 \\
& \text{subject to } \|D^{(j)}\|_2 = 1, \, \forall j  \nonumber \\
& \text{and }  \| w^{(i)} \|_0 \leq \gamma, \, \forall i \, . \nonumber
\end{align}
We again used ODL~\citep{mairal2009online} to learn the dictionary, but this time solved the sparse weights with Orthogonal Matching Pursuit (OMP)~\citep{pati1993orthogonal}. In our experiments, we cross-validated the sparsity parameter $\gamma$ with $\gamma \in \{1, 5, 10, 15\}$.

\subsubsection{Gain-Shape Vector Quantization (GSVQ)}

The idea is to represent a vector by separating its gain (euclidean norm) from its shape (orientation)~\citep{gersho2012vector}. GSVQ has a similar formulation to OMP where $\gamma=1$. Specifically, it computes $k = \arg \max_j | D^{(j)} \cdot x^{(i)} |$, the atom index that is most correlated with $x^{(i)}$, then sets $w^{(i)}_k = D^{(k)} \cdot x^{(i)}$ and $w^{(i)}_j = 0$ for $j \neq k$. Using these fixed weight vectors, it is then straightforward to find the locally optimal dictionary $D$ in \eqref{eq:dl_omp} using an iterative procedure as in KMeans.

\subsubsection{Normalized KMeans (NKM)}

\citet{coates2011analysis} showed that the centroids learned by a standard KMeans algorithm make good dictionary atoms. We thus clustered the patches and used the normalized centroids as dictionary atoms.

\subsubsection{Randomly Sampled Patches (RP)}

Here, we used the heuristic proposed by~\citet{coates2011importance} to populate the dictionary. We uniformly sampled $d$ patches from the dataset and used the normalized vectors as the dictionary atoms.

\subsubsection{Random Dictionary (R)}

It has been shown previously that completely random weights can achieve surprisingly good results~\citep{coates2011importance, saxe2011random}. Therefore, we also tested \textit{learning} the dictionary by sampling $d$ times the uniform distribution $U\left( [0,1]^n \right)$ and used the normalized vectors as dictionary atoms.

\subsection{Feature Coding}

Executing any dictionary learning approaches presented in the previous section gives a dictionary $D$ representing the latent structure of the data. To actually extract features from them, we use an \textit{encoder} that maps an observation $x^{(i)}$ to its feature representation $f^{(i)}$.

\subsubsection{Sparse Coding (SC)}

The first approach is based on the sparse coding formulation of \eqref{eq:dl_sc}. Given a dictionary learned with any of section~\ref{ssec:dl} methods (not necessarily SC), we solve \eqref{eq:dl_sc} for the sparse weights assuming a fixed $D$:
\begin{align}
\label{eq:fc_sc}
w^{(i)} &= \arg \min_{w^{(i)}} \| D w^{(i)} - x^{(i)} \|_2^2 + \lambda \| w^{(i)} \|_1, \, \forall i \, ,
\end{align}
using LARS~\citep{efron2004least}. It is important to note that $\lambda$ (also $\gamma$ and $\tau$) may have a different value during feature coding than dictionary learning. We then apply \textit{polarity splitting}~\citep{coates2011importance}, that is, we split the positive weights from the negative ones:
\begin{align*}
f^{(i)}_j = \max\{w^{(i)}_j, 0\} \\
f^{(i)}_{j+d} = \max\{-w^{(i)}_j, 0\}
\end{align*}
This technique allows the classifier to model positive and negative weights differently, thus improving its flexibility~\citep{coates2011importance}. In our experiments, we cross-validated with $\lambda \in \{0.5, 1, 1.5, 2\}$.

\subsubsection{Masked Sparse Coding (mSC)}
With this approach, we explicitly deal with masked-out entries arising from either the preservation of the aspect ratio during grasp rectangle rescaling or noisy Kinect depth sensor. Let $m^{(i)} \in \{0,1\}^n$ be the mask vector of observation $x^{(i)}$ where $m^{(i)}_j = 0$ implies that $x^{(i)}_j = 0$ indicates a masked entry (similarly, $m^{(i)}_j = 1$ indicates that $x^{(i)}_j \gtreqless 0$ is not a masked entry). We then remove the penalty induced by the masked entries in \eqref{eq:fc_sc} giving the following formulation, $\forall i $:
\begin{align}
\label{eq:fc_msc}
w^{(i)} &= \arg \min_{w^{(i)}} \| \text{diag}(m) (D w^{(i)} - x^{(i)}) \|_2^2 + \lambda \| w^{(i)} \|_1
\end{align}
To solve \eqref{eq:fc_msc}, we again used LARS~\citep{efron2004least}, this time with the mask. As in SC, we apply polarity splitting. In our experiments, we cross-validated with $\lambda \in \{1, 2, 3, 4\}$.

\subsubsection{Orthogonal Matching Pursuit (OMP)}

This approach is based on the formulation of \eqref{eq:dl_omp}. Assuming a fixed dictionary, we applied OMP~\citep{pati1993orthogonal} to solve for the sparse weights:
\begin{align}
\label{eq:fc_omp}
w^{(i)} &= \min_{w^{(i)}} \| D w^{(i)} - x^{(i)} \|_2^2 \\
& \text{subject to } \| w^{(i)} \|_0 \leq \gamma, \, \forall i \nonumber \, .
\end{align}
We again used polarity splitting as in SC. We cross-validated with $\gamma \in \{1, 5, 10, 15\}$.

\subsubsection{Masked Orthogonal Matching Pursuit (mOMP)}

Similarly to mSC, we remove the penalty of the masked entries from \eqref{eq:fc_omp} giving the following formulation
\begin{align}
w^{(i)} &= \min_{w^{(i)}} \| \text{diag}(m) (D w^{(i)} - x^{(i)}) \|_2^2 \\
& \text{subject to } \| w^{(i)} \|_0 \leq \gamma, \, \forall i \nonumber \, ,
\end{align}
which we solved using OMP~\citep{pati1993orthogonal} but this time with the mask. We used polarity splitting and cross-validated with $\gamma \in \{1, 5, 10, 15\}$.

\subsubsection{Soft-Thresholding (ST)}

Soft-Thresholding~\citep{donoho1995adapting} (also known as marginal regression~\citep{genovese2012comparison}) is a fast alternative to finding the optimal solution of \eqref{eq:fc_sc}. It is based on the (strong) hypothesis that all weights $w^{(i)}_j$ are independent. This enables solving \eqref{eq:fc_sc} for the sparse weights marginally (each $w^{(i)}_j$ individually) thus giving a simple analytical solution:
\begin{align}
w_j^{(i)} = \text{sign}(D^{(j)} \cdot x^{(i)}) \cdot \max \left\{ 0, | D^{(j)} \cdot x^{(i)} | - \tau \right\} \, ,
\end{align}
where $\tau$ is the sparsity parameter.  Similar to SC and OMP, we applied polarity splitting on $w^{(i)}$. We cross-validated with $\tau \in \{0.5, 1, 1.5, 2\}$.

\subsubsection{Natural (N)}

Finally, we define a \textit{natural} encoder as whichever approach was used for solving the sparse weights during dictionary learning. For instance, the natural encoder of SC dictionary learning is SC feature coding with the same sparsity parameter $\lambda$. Similarly, the natural encoder of OMP dictionary learning is OMP feature coding with the same $\gamma$. We used a different approach for the other dictionary learning algorithms, since they do not require a sparsity parameter. Specifically, for GSVQ we used OMP with $\gamma = 1$. For R and RP we used ST with $\tau=0$ which corresponds to a random linear projection. Finally, for NKM we did not normalize the centroids (as in standard KMeans) and used the KMeans-Tri feature coding of~\citet{coates2011analysis}.

\subsection{Feature Extraction Process}
\label{ssec:feature_extraction_process}

\begin{figure*}[t]
\centering
\includegraphics[width=\textwidth]{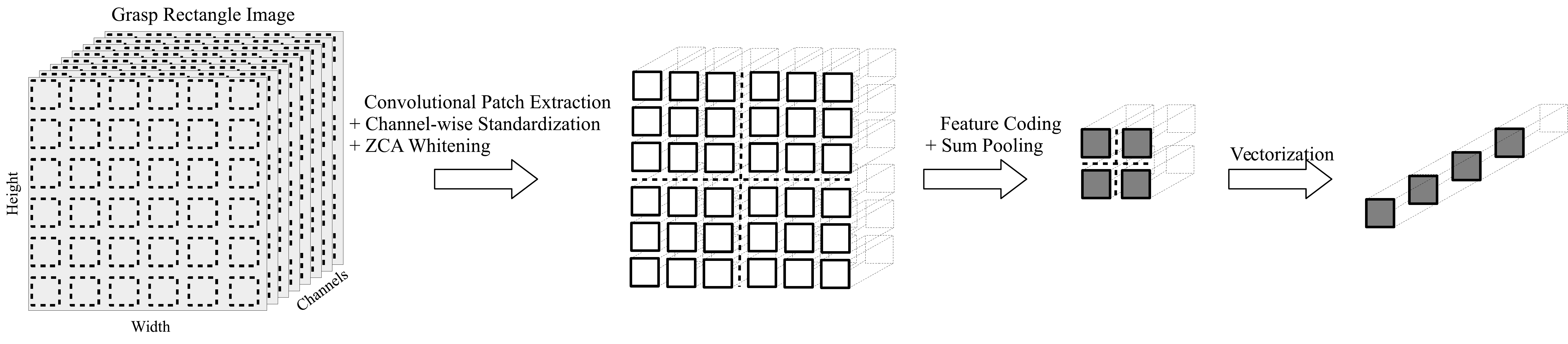}
\caption{Feature extraction process. Left: a batch of $6 \times 6 \times 8$ patches $x^{(i)}$ are extracted in a convolutional way (with a stride of one pixel) from the image. The patches are then channel respective standardized and ZCA whitened. Center-Left: each patch $x^{(i)}$ is mapped to its feature representation $f^{(i)}$ given a dictionary $D$ and a choice of encoder. Center-Right: a four quadrants sum pooling is applied on feature vectors $f^{(i)}$. Right: all quadrant pooled weights are concatenated into a single vector. The final vector on the right is used as input to the SVM.}
\label{fig:pipeline_feature_coding}
\end{figure*}

Given a learned dictionary $D$ and a choice of encoder, a patch vector $x^{(i)}$ can now be transformed into its feature representation $f^{(i)}$. Here we describe how to transform a full grasp rectangle image into a feature representation usable by the classifier for grasp recognition. Our feature extraction process follows the spatial pyramid matching framework proposed by~\citet{yang2009linear}. The entire process is shown in Fig.~\ref{fig:pipeline_feature_coding}. Specifically, we extract a batch of $6 \times 6 \times 8$ patches $x^{(i)}$ in a convolutional way with a stride of one pixel. These patches are channel-wise standardized and ZCA whitened. Then, we map each patch to its feature representation $f^{(i)}$ using the dictionary $D$ and the encoder. After, we divide the image into four quadrants and perform sum pooling over the feature vectors $f^{(i)}$ of each quadrant respectively. Finally, the pooled feature vectors from all quadrants are concatenated into a single vector that is used as input to the classifier.

\subsection{Grasp Recognition Classifier}

For classification, we optimized a $\ell_2$-linear SVM using a standard L-BFGS solver from Schmidt's \textit{minFunc} toolbox~\citep{minFunc}. We cross-validated the regularization parameter $C$ with $C \in \{1, 10, 100, 1000\}$.

\subsection{Grasp Detection Regressor}

Although performing grasp recognition is straightforward with a SVM because it is a binary classification problem, grasp detection is more cumbersome. Directly predicting the best $5$-dimensional grasp rectangle from the very high-dimensional inputs ($640 \times 480 \times 4$ Kinect images) would not be realistic with our current framework. A naive application of the feature extraction process described in section~\ref{ssec:feature_extraction_process} on an entire Kinect image would extract an unreasonably high-dimensional feature vector with little discriminative power. We instead opted for a standard grid-search in grasp rectangle space. We first performed background removal and identified the smallest region containing the object. Then, we extracted, in a convolutional way with a stride of 10 pixels, grasp rectangles from the image with varying sizes and orientations. We varied the size of the rectangle from 10 pixels to 90 pixels with a stride of 10 pixels, and varied the orientation from 0 degree to 180 degrees with a stride of 15 degrees. For each of these grasp rectangle images, we performed feature extraction and inputed them to the SVM. The rectangle having the highest classification score was chosen as the candidate grasp.

\section{Experimental Framework}
\label{sec:experimental_framework}

\subsection{Cornell Dataset}

The Cornell Grasping Dataset~\citep{jiang2011efficient} contains 885 RGBD images of 240 distinct objects (available at \url{http://pr.cs.cornell.edu/deepgrasping/}). Each image has multiple positively- and negatively-labeled grasp rectangles, specifically selected for parallel plate grippers. The labeled rectangles are varied in terms of size, orientation and position, but are by no means exhaustive of every grasp scenarios (some image have graspable regions that are not labeled).

\subsection{Grasp Recognition Experiments}

For grasp recognition, we performed the following three experiments:

\subsubsection{Grasp Recognition Evaluation}

Previous works on the Cornell dataset reported their results using a 5-fold cross-validation \citep{jiang2011efficient, lenz2015deep}. They optimized for the hyper-parameters using a separate set of grasp examples (which we call the validation set). However, exactly comparing our results to theirs is impossible because they did not report which examples they selected for validation. Therefore, we instead report our recognition accuracies using a 5-5 folds nested cross-validation. The advantages are that this removes the need to specify the validation set and reduces the bias induced by choosing which examples to put in it. Moreover, our results are still comparable to the previous ones and, most importantly, allows future works on the dataset to report results based on the same evaluation framework.

\subsubsection{Varying the Dictionary Size}

The number $d$ of atoms has a direct influence on the performance of the classifier. On one side, a too small dictionary $D$ does not capture enough structure to correctly represent the raw data. On the contrary, a too big dictionary $D$ contains noisy atoms which are never activated in the sparse weight vectors $w^{(i)}$. These unused atoms slow down the weight extraction process and make the resulting feature vector $f^{(i)}$ unnecessarily long and noisy. We therefore evaluate the correlation between the recognition accuracy and the dictionary size $d$ to later guide our choice of it during grasp detection.

\subsubsection{ZCA Whitening}

Previous works on dictionary learning applied to RGB object recognition have already shown that whitening improves the accuracy of the classifier~\citep{coates2011analysis, coates2011importance}. We therefore wanted to validate that it is still the case when applying DLSR approaches for grasp recognition in RGBD images.

\subsection{Grasp Detection Experiments}

For grasp detection, we performed the following two experiments:

\subsubsection{Grasp Detection Evaluation}

To evaluate the quality of a grasp candidate, we used the \textit{rectangle metric} as in~\citep{jiang2011efficient, lenz2015deep}. Specifically, if the rectangle metric of any of the ground truth rectangles with the candidate is positive, the regression is a success. In more detail, the metric is positive if: \textit{1)} the candidate orientation is within $30^{\circ}$ of the ground truth rectangle, and \textit{2)} the Jaccard index between the candidate and the ground truth is greater than $25\%$, where the Jaccard index between two rectangles $R_1$ and $R_2$ is defined as:
\begin{align}
J\left(R_1, R_2\right) &= \frac{\text{area}(R_1 \cap R_2)}{\text{area}(R_1 \cup R_2)} \, .
\end{align}

Unlike grasp recognition, we performed a standard 5 folds cross-validation for the detection problem. We did not optimize for the hyper-parameters, but instead used those that were the most often selected in the nested (second layer) cross-validation during grasp recognition. We also used two learning scenarios~\citep{lenz2015deep}:
\begin{itemize}
\item Image-wise splitting: where we split the images randomly.
\item Object-wise splitting: where we split the objects randomly, gathering all the image of the object in the same fold.
\end{itemize}

Image-wise splitting studies the ability to generalize to new positions and orientations of an object that has already been seen. Object-wise splitting examine the capability to generalize to novel, unseen objects. While the first scenario is more suitable in an industrial context, because the set of objects is known beforehand, the second one is more difficult but also more realistic. Since training on all possible objects is almost impossible, this help asserting the viability of the approach to perform everyday grasping.

\subsubsection{Self-Taught Learning}

One of the most appealing advantage of unsupervised feature learning is the possibility of using unlabeled data for learning the dictionary $D$. We therefore evaluated the approaches on the problem of \textit{self-taught learning}~\citep{raina2007self}. Specifically, we randomly subsampled 30 objects from the Washington RGBD dataset, which also contains Kinect images of everyday objects~\citep{lai2011large}. We then randomly selected 25 images per object, giving 750 images in total. From these images, we further extracted 100,000 $6 \times 6 \times 8$ patches and added them to the 100,000 patches extracted from the Cornell dataset images. We then learned the dictionary $D$ using the patches from both datasets. This test examined the capacity of learning useful features from images taken from another distribution that will never be tested on.

\section{Experimental Results}
\label{sec:experimental_results}

\subsection{Grasp Recognition Results}

\subsubsection{Grasp Recognition Evaluation}

\begin{table}[t]
\caption{Cross-validation results of all combinations of dictionary learning and feature coding, for Cornell dataset. Numbers are grasp recognition accuracies, in percent (\%), from 5-5 folds nested cross validation where hyper-parameter maximization is performed on the nested folds.}
\begin{center}
\begin{tabular}{c|cccccc}
\hline \hline
\multicolumn{1}{c|}{Learning \: \rotatebox{90}{Features}} & \rotatebox{90}{SC} & \rotatebox{90}{mSC} & \rotatebox{90}{OMP}  & \rotatebox{90}{mOMP} & \rotatebox{90}{ST} &  \rotatebox{90}{Natural} \\
\hline 
SC & 96.63 & 96.74 & 96.61 & 96.61 & 96.73 & 96.65 \\ 
OMP & 96.68 & 96.71 & 96.60 & 96.69 & 96.50 & 96.58  \\ 
GSVQ & 96.72 & 96.50 & 96.66  & 96.70  & 96.50 & 95.65  \\ 
NKM & 96.86 & 96.64 & 96.58 & 96.64 &  96.42 & 96.52 \\ 
RP & 96.43 & 96.51 & 96.35 & 96.20 & 96.28 & 96.37 \\ 
R & 95.84 & 95.52 & 95.34 & 95.15 & 95.78 & 95.90 \\ 
\hline \hline
\end{tabular}
\end{center}
\label{tab:comparison_recognition_results}
\end{table}

The nested cross-validation accuracies (in \%) for grasp recognition using a dictionary of $d$=300 atoms are reported in Table~\ref{tab:comparison_recognition_results}. The best dictionary learning + encoder combination was NKM-SC, which reached an accuracy of $96.86\%$, while the lowest accuracy is obtained with R-mOMP at $95.15\%$. As a comparison, previous approaches from~\citet{jiang2011efficient}, who used a cascade of multi-layer perceptrons, achieved $89.6\%$ and~\citet{lenz2015deep}, who used a ImageNet pre-trained convolutional neural network (CNN), had $93.7\%$. These results suggests that even though neural network-based models are nowadays most popular way to solve vision-related problems, DLSR is still a viable approach because \textit{1)} it obtained the highest accuracy on Cornell recognition task and \textit{2)} the training cost is significantly smaller than CNN in both the computation time and the training dataset size. 

Table~\ref{tab:comparison_recognition_results} shows that, apart from method R, any DLSR combination arrives at similar performances (accuracies vary by no more than $1\%$). This suggests that selecting a dictionary learning or a feature coding approach may only be based on the practical consideration that it requires no hyper-parameter tuning. For instance, GSVQ, NKM and RP dictionary learning approaches could be considered before SC and OMP due to their hyper-parameter free nature (SC has $\lambda$ and OMP has $\gamma$). Similarly, NKM, GSVQ and RP natural features may be taken in consideration before all other feature encoding because they have no hyper-parameters.

\begin{figure}[t]
    \centering
    \begin{minipage}{0.3\textwidth}
        \centering
        \includegraphics[width=\textwidth]{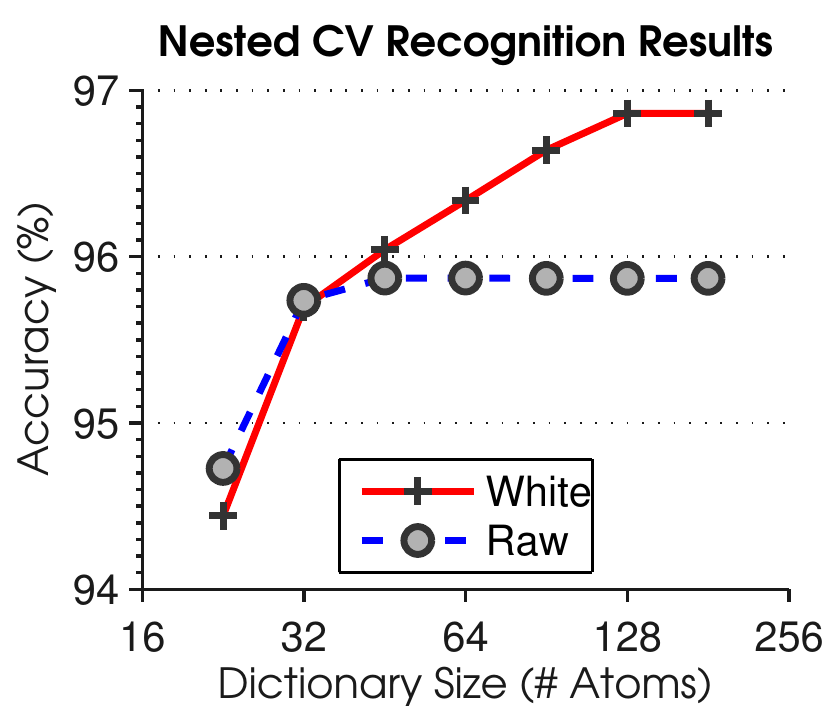}
        {\scriptsize NKM-SC}
    \end{minipage}%
    \begin{minipage}{0.3\textwidth}
        \centering
        \includegraphics[width=\textwidth]{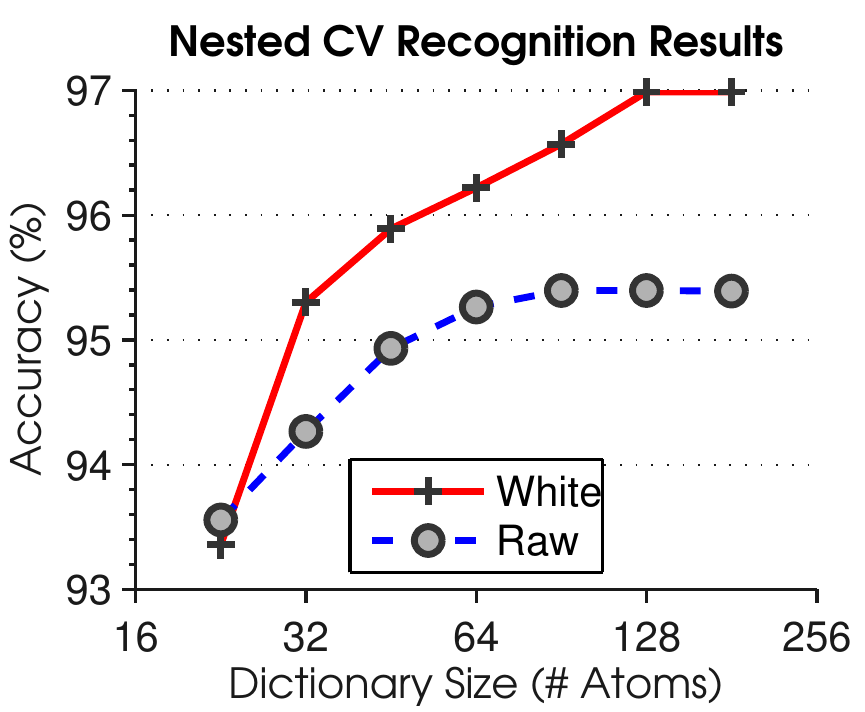}
        {\scriptsize GSVQ-ST}
    \end{minipage}%
    \begin{minipage}{0.3\textwidth}
        \centering
        \includegraphics[width=\textwidth]{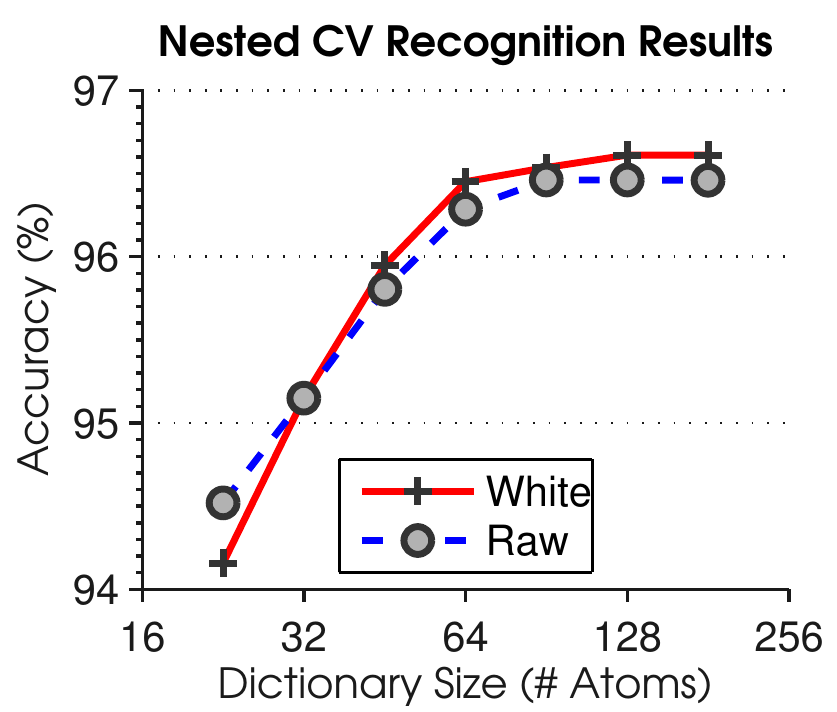}
        {\scriptsize OMP-Natural}
    \end{minipage}%
    
    \begin{minipage}{0.3\textwidth}
        \centering
        \includegraphics[width=\textwidth]{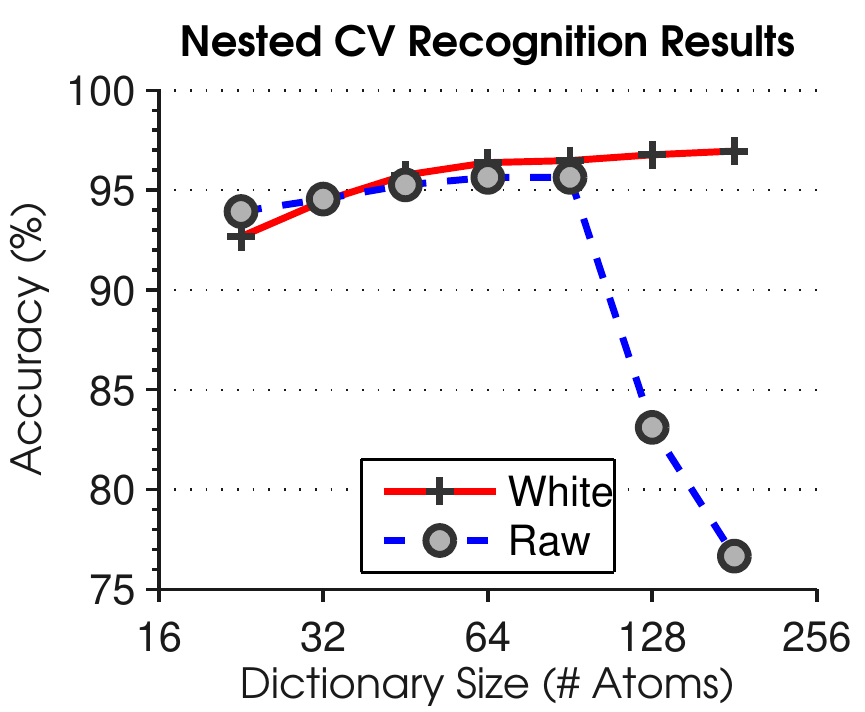}
        {\scriptsize NKM-Natural}
    \end{minipage}%
    \begin{minipage}{0.3\textwidth}
        \centering
        \includegraphics[width=\textwidth]{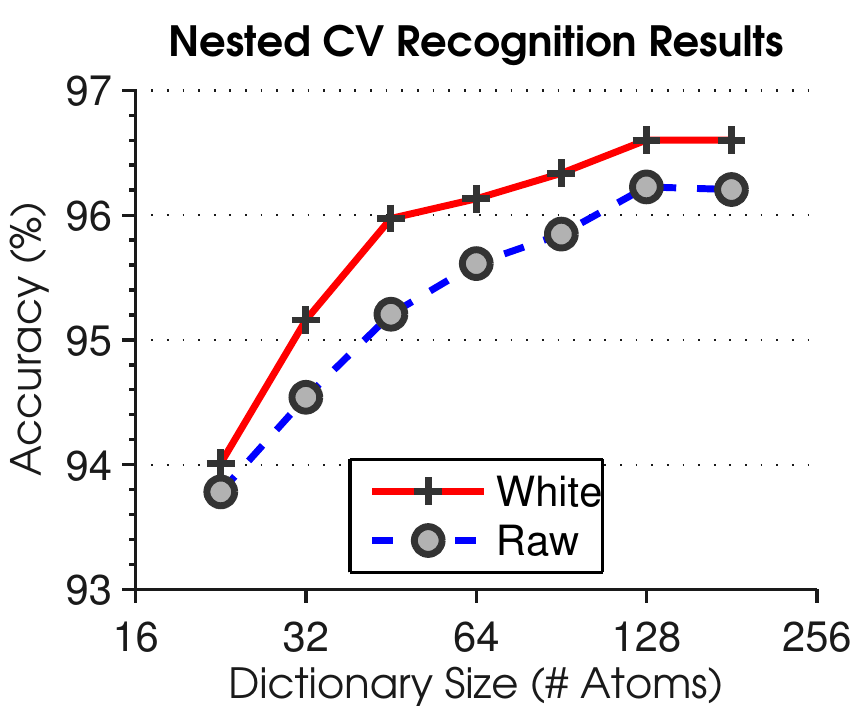}
        {\scriptsize RP-Natural}
    \end{minipage}%
    \caption{The effects of whitening and the dictionary size on recognition accuracies. Whitening improves the performance and may degrade it when not used. Increasing the dictionary size improves the results up to a limit (plateau) where no more boost is possible.} 
    \label{fig:dictiosize}
\end{figure}

The best overall encoders across the dictionary learning approaches are SC and mSC. The $\ell_1$ optimization~\eqref{eq:fc_sc}, along with its masked version~\eqref{eq:fc_msc}, appear to extract the most robust features. However, these encoders were the most time consuming of all, due to the tedious optimization needed for solving the sparse weights $w^{(i)}$. For real-time scenarios, these encoders would be too cumbersome to use. We thus select GSVQ-ST, NKM-Natural and RP-Natural as the three most appealing combinations and use them for the next experiments. NKM-Natural and RP-Natural are both hyper-parameter free, have a fast encoder and dictionary learning algorithm. Even though GSVQ-ST defines the $\lambda$ hyper-parameter, it is a fast encoder which shows good recognition accuracies. For more thorough evaluations, we also keep the top performer NKM-SC and OMP-Natural for its fast greedy optimization.

\subsubsection{The Effects of Whitening and Dictionary Size}

\begin{figure}[t]
    \centering
    \begin{minipage}{0.3\linewidth}
        \centering
        \includegraphics[width=\linewidth]{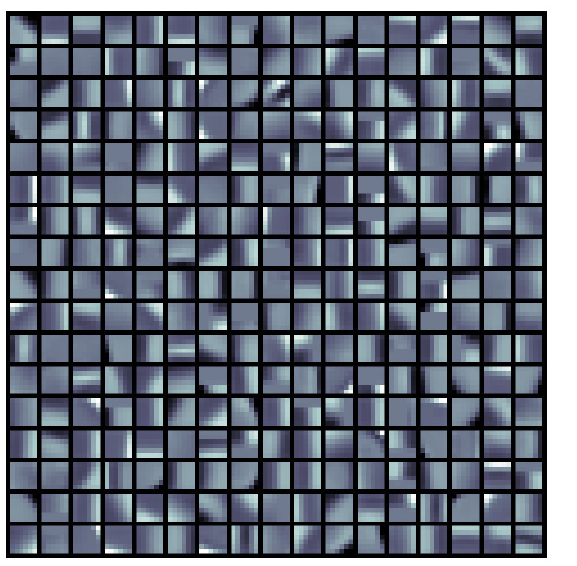}
        {\scriptsize $K$}
    \end{minipage}%
    \begin{minipage}{0.3\linewidth}
        \centering
        \includegraphics[width=\linewidth]{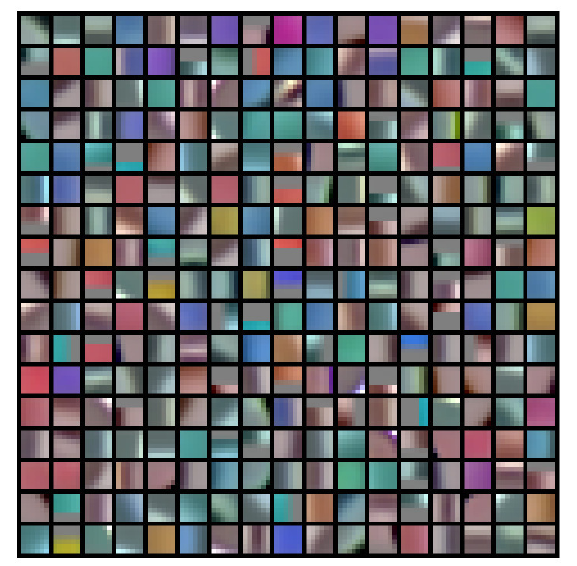}
        {\scriptsize $RGB$}
    \end{minipage}%
    
    \begin{minipage}{0.3\linewidth}
        \centering
        \includegraphics[width=\linewidth]{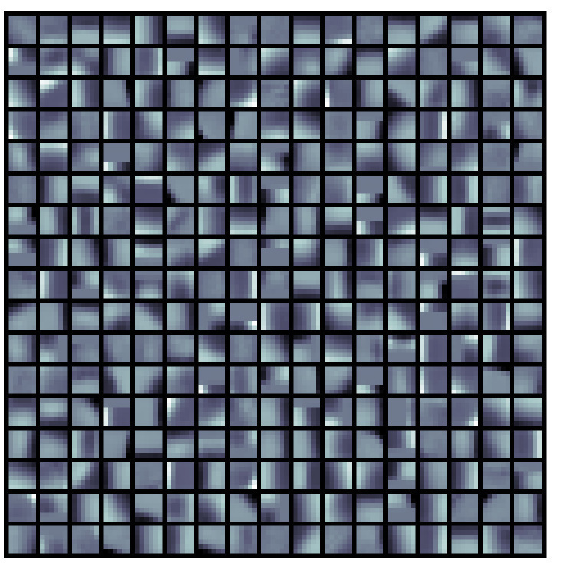}
        {\scriptsize $D$}
    \end{minipage}%
    \begin{minipage}{0.3\linewidth}
        \centering
        \includegraphics[width=\linewidth]{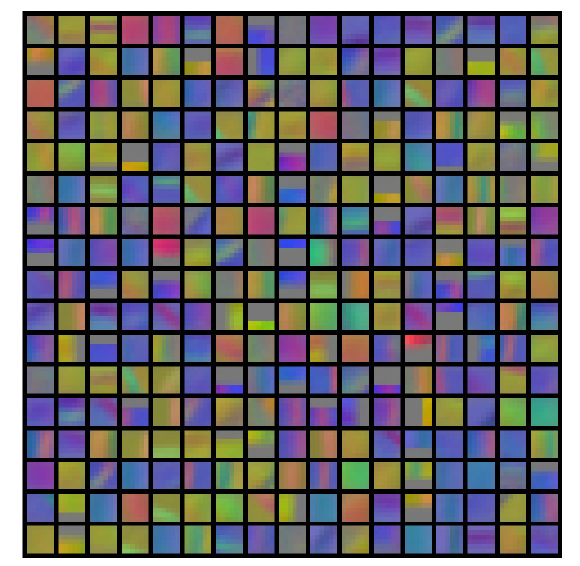}
        \vspace{-3pt}
        {\scriptsize  $N_x N_y N_z$}
    \end{minipage}%
    \caption{A dictionary $D$ of 300 atoms (each square is an atom $D^{(j)}$) learned using the Cornell dataset shown in four distinct parts: K (gray), RGB, D (depth) and $N_x N_y N_z$ (depth normals). No whitening is performed, and we clearly see the absence of localized and oriented Gabor-like filters.}
    \label{fig:dictionary_no_white}
\end{figure}

The effects of whitening along with varying the dictionary size $d$ are displayed in Fig.~\ref{fig:dictiosize}. Whitening improves performance, and can degrade results when not used, as seen with NKM-Natural. Increasing the dictionary size $d$ improves the results up to a limit where no additional gain is possible. The plateau indicates that dictionary learning is unable to learn new useful features, thus reaching a limit in its representative capability. We can see visually the impact of whitening in Fig.~\ref{fig:dictionary_no_white}. The dictionary learned with raw images clearly shows a lack of Gabor-like filters. This is because whitening removes redundant information in the inputs, hence learning more discriminative features. 

\subsection{Grasp Detection Results}

\subsubsection{Grasp Detection Evaluation}

The cross-validation accuracies (in \%) for grasp detection of the five selected approaches using a dictionary of $d=300$ atoms are reported in Table~\ref{tab:detection_results}. For image-wise and object split respectively, the best accuracies are obtained by NKM-Natural with $89.40\%$ and GSVQ-ST with $88.79\%$. As comparison,~\citet{lenz2015deep} obtained $73.9\%$ and $75.6\%$ with a cascade of multi-layer perceptrons while~\citet{redmonRealtimeCnnGrasp} achieved $88.0\%$ and $87.1\%$ with a CNN.

Even though CNN is a powerful approach (currently state-of-the-art in several vision-related problems), DLSR has one advantage CNN has not. Due to the relative small amount of data in Cornell dataset, Redmon \textit{et al.} pre-trained their CNN on ImageNet containing RGB images, replaced blue with depth channel to make it compatible with RGBD images (giving RGD images), and performed a final fine-tuning on Cornell images. Since blue channel-related low level features are unlikely extracting useful information from depth, such a pre-training approach is clearly sub-optimal. Directly training the CNN on Cornell dataset would require gathering a large quantity of additional images, at a substantial cost. In contrast, DLSR can be directly trained on Cornell RGBD images despite its small size, and this makes it more advantageous in this manner.

\begin{table}[t]
\caption{Cross validation detection results for the Cornell dataset. See text for a discussion on computing time.}
\begin{center}
\begin{tabular}{c|cc}
\hline \hline
\multirow{2}{*}{Algorithm} & \multicolumn{2}{c}{Detection Accuracy (\%)} \\
& Image-wise Split & Object-wise Split \\
\hline
\citet{jiang2011efficient} & 60.5 & 58.3 \\
\citet{lenz2015deep} & 73.9 & 75.6 \\
\citet{redmonRealtimeCnnGrasp} & 88.0 & 87.1 \\
\hline
NKM-SC & 88.67 & 88.07 \\
GSVQ-ST & 88.72 & \textbf{88.79} \\
OMP-Natural &  89.34 & 88.56 \\
NKM-Natural & \textbf{89.40} & 88.17 \\
RP-Natural & 87.70 & 86.61 \\
\hline \hline
\end{tabular}
\end{center}
\label{tab:detection_results}
\end{table}

\subsubsection{Self-Taught Learning}

\begin{table}[t]
\caption{Cross validation detection results for the Cornell dataset using patches from both Cornell and Washington datasets.}
\begin{center}
\begin{tabular}{c|cc}
\hline \hline
\multirow{2}{*}{Algorithm} & \multicolumn{2}{c}{Detection Accuracy (\%)} \\
& Standard & Self-Taught \\
\hline
NKM-SC & 88.07 & 88.85 \\
GSVQ-ST & 88.79 & 87.76 \\
OMP-Natural & 88.56 & 88.18 \\
NKM-Natural &  88.17 & 88.53 \\
RP-Natural & 86.61 & 86.12 \\
\hline \hline
\end{tabular}
\end{center}
\label{tab:detection_results_selftaught}
\end{table}

The cross-validation accuracies (in \%) for grasp detection of the five selected approaches using a dictionary of 300 atoms and patches from both Cornell and Washington datasets are reported in Table~\ref{tab:detection_results_selftaught}. Even tough we see both a performance improvement (NKM-SC and NKM-Natural) and decrease (GSVQ-ST, OMP-Natural and RP-Natural), the variations are small (around $0.5\%$) and not significant. This suggests that the dictionary learning approaches were able to extract all the necessary features from the Cornell images, and the additional features extracted from the Washington dataset were not helpful. This is further reflected in Fig.~\ref{fig:dictionary} by the presence of some weakly structured atoms (blank squares). Indeed, a dictionary learning approach normally use all available atoms to well represent highly structured data. Here, it could achieve that by using only a subset of the atoms, thus showing that it did not need additional patches to learned all the relevant features.

\section{Discussion}
\label{sec:discussion}

To obtain Table~\ref{tab:detection_results} high accuracy results in detection, we had to pay a computational price. Even though feature coding approaches have reasonable low computational complexity (SC is $O(d^3+nd^2)$, OMP is $ O(\gamma  n  d) $ and ST is $O(n  d)$), the exhaustive grid search in grasp rectangle space is computationally demanding. This translates into several minutes to complete a detection which is higher than Redmon \textit{et al.}'s CNN with $76$ ms per image. However, since we used a standard CPU and they used a high-end GPU, a DLSR GPU implementation would make a fairer time computation comparison. While implementing ST would be simple, this is not straightforward for SC and OMP due to their recursive nature. A possible avenue for a useful GPU implementation may be to parallelize grid search by exploiting grasp rectangle candidate independence, i.e. by extracting and scoring all candidates in parallel. However, grid search in grasp rectangle space being more cumbersome than spatial convolutions, the computational time of such a parallelization would still be higher than CNN.

While DLSR is fairly slow during detection, the training phase is significantly faster than CNN. Training a CNN (as the one used by~\citet{redmonRealtimeCnnGrasp}) take several days with parallel high-end GPUs, and fine-tuning on Cornell dataset takes several hours. In comparison, it took approximatively ten minutes to train our $300$ atoms dictionaries on a standard CPU. While fast training phase is irrelevant in real-time test scenarios, it could be useful to train a CNN directly on RGBD images. \citet{redmonRealtimeCnnGrasp} pre-training on ImageNet could be avoided by greedily stacking dictionaries learned on RGBD images, as previously proposed by~\citet{bengio2007greedy} with auto-encoders in the context of RGB images. Such a DLSR and CNN combination would bring the best of both approaches, in which DLSR would improve training while CNN would provide fast detection. We intend to investigate this avenue in future works.

Even though it achieved the lowest detection accuracies, the RP-Natural combination is appealing because training the dictionary is instantaneous, feature coding requires only a matrix multiplication, and there is no hyper-parameters. Due to its simplicity, integrating the approach to a grasp localization system in its early deployment phase is straightforward and can give a good glimpse of the overall system performance in later deployment phases. One interesting avenue for future work is to understand the reason why input decorrelation allows randomly sampled patches to make such good dictionaries. For instance, is linear independence sufficient, or better dictionaries could be recovered with non-linear independence? These are several avenues worth investigating.

\section{Conclusion}
\label{sec:conclusion}

A perception system that determines good grasping positions from Microsoft Kinect RGBD images is a key element toward automating the grasping of ordinary objects. In this paper, we proposed a DLSR framework to recognize and detect grasp rectangles on images of object to be held by two-plates parallel grippers. Our comparative study of various dictionary learning and feature coding approach combinations on Cornell dataset have shown that the proposed DLSR framework outperformed previous neural network-based approaches. As opposed to CNN, the best DLSR combination obtained a greater accuracy in both grasp recognition and detection task despite training only on small amount of images. In addition to having a substantially fast training phase, DLSR can inherently deal with masked-out entries in noisy depth maps and do not rely on sophisticated regularization terms. As discussed in section~\ref{sec:discussion}, exploiting DLSR fast training phase may be a suitable research avenue for future work. Stacking dictionaries learned on RGBD images to pre-train a CNN would bring the best of both approaches, in which DLSR improve training while CNN provide fast detection.

\setlength{\bibsep}{4pt plus 0.3ex}

\bibliographystyle{apalike}
{\small
\bibliography{bibliography}

\begin{thebibliography}{}

\bibitem[Aharon et~al., 2006]{aharon2006img}
Aharon, M., Elad, M., and Bruckstein, A. (2006).
\newblock {KSVD}: An algorithm for designing overcomplete dictionaries for
  sparse sepresentation.
\newblock {\em Signal Processing, Transactions on}, 54(11):4311--4322.

\bibitem[Bengio et~al., 2007]{bengio2007greedy}
Bengio, Y., Lamblin, P., Popovici, D., Larochelle, H., et~al. (2007).
\newblock Greedy layer-wise training of deep networks.
\newblock {\em Advances in neural information processing systems}, 19:153.

\bibitem[Blum et~al., 2012]{blum2012learned}
Blum, M., Springenberg, J.~T., W{\"u}lfing, J., and Riedmiller, M. (2012).
\newblock A learned feature descriptor for object recognition in {RGB-D} data.
\newblock In {\em ICRA}, pages 1298--1303.

\bibitem[Bo et~al., 2013]{bo2013unsupervised}
Bo, L., Ren, X., and Fox, D. (2013).
\newblock Unsupervised feature learning for {RGB-D} based object recognition.
\newblock In {\em Experimental Robotics}, pages 387--402.

\bibitem[Coates and Ng, 2011]{coates2011importance}
Coates, A. and Ng, A.~Y. (2011).
\newblock The importance of encoding versus training with sparse coding and
  vector quantization.
\newblock In {\em ICML}, pages 921--928.

\bibitem[Coates et~al., 2011]{coates2011analysis}
Coates, A., Ng, A.~Y., and Lee, H. (2011).
\newblock An analysis of single-layer networks in unsupervised feature
  learning.
\newblock In {\em AISTATS}, pages 215--223.

\bibitem[Detry et~al., 2013]{detry2013learning}
Detry, R., Ek, C.~H., Madry, M., and Kragic, D. (2013).
\newblock Learning a dictionary of prototypical grasp-predicting parts from
  grasping experience.
\newblock In {\em ICRA}, pages 601--608.

\bibitem[Donoho and Johnstone, 1995]{donoho1995adapting}
Donoho, D.~L. and Johnstone, I.~M. (1995).
\newblock Adapting to unknown smoothness via wavelet shrinkage.
\newblock {\em Journal of the american statistical association},
  90(432):1200--1224.

\bibitem[Efron et~al., 2004]{efron2004least}
Efron, B., Hastie, T., Johnstone, I., Tibshirani, R., et~al. (2004).
\newblock Least angle regression.
\newblock {\em The Annals of statistics}, 32(2):407--499.

\bibitem[Elad and Aharon, 2006]{elad2006image}
Elad, M. and Aharon, M. (2006).
\newblock Image denoising via sparse and redundant representations over learned
  dictionaries.
\newblock {\em Image Processing, Transactions on}, 15(12):3736--3745.

\bibitem[Genovese et~al., 2012]{genovese2012comparison}
Genovese, C.~R., Jin, J., Wasserman, L., and Yao, Z. (2012).
\newblock A comparison of the lasso and marginal regression.
\newblock {\em JMLR}, 13(1):2107--2143.

\bibitem[Gersho and Gray, 2012]{gersho2012vector}
Gersho, A. and Gray, R.~M. (2012).
\newblock {\em Vector quantization and signal compression}, volume 159.
\newblock Springer Science \& Business Media.

\bibitem[Goldfeder et~al., 2007]{goldfeder2007grasp}
Goldfeder, C., Allen, P.~K., Lackner, C., and Pelossof, R. (2007).
\newblock Grasp planning via decomposition trees.
\newblock In {\em ICRA}, pages 4679--4684.

\bibitem[Gupta et~al., 2014]{gupta2014learning}
Gupta, S., Girshick, R., Arbel{\'a}ez, P., and Malik, J. (2014).
\newblock Learning rich features from {RGB-D} images for object detection and
  segmentation.
\newblock In {\em ECCV}, pages 345--360.

\bibitem[Hyv{\"a}rinen et~al., 2004]{hyvarinen2004independent}
Hyv{\"a}rinen, A., Karhunen, J., and Oja, E. (2004).
\newblock {\em Independent component analysis}, volume~46.
\newblock John Wiley \& Sons.

\bibitem[Jiang et~al., 2011]{jiang2011efficient}
Jiang, Y., Moseson, S., and Saxena, A. (2011).
\newblock Efficient grasping from {RGBD} images: Learning using a new rectangle
  representation.
\newblock In {\em ICRA}, pages 3304--3311.

\bibitem[Lai et~al., 2011]{lai2011large}
Lai, K., Bo, L., Ren, X., and Fox, D. (2011).
\newblock A large-scale hierarchical multi-view {RGB-D} object dataset.
\newblock In {\em ICRA}, pages 1817--1824.

\bibitem[Lalibert{\'e} et~al., 2002]{laliberte2002underactuation}
Lalibert{\'e}, T., Birglen, L., and Gosselin, C. (2002).
\newblock Underactuation in robotic grasping hands.
\newblock {\em Machine Intelligence \& Robotic Control}, 4(3):1--11.

\bibitem[Lazebnik et~al., 2006]{lazebnik2006beyond}
Lazebnik, S., Schmid, C., and Ponce, J. (2006).
\newblock Beyond bags of features: Spatial pyramid matching for recognizing
  natural scene categories.
\newblock In {\em CVPR}, volume~2, pages 2169--2178.

\bibitem[Le et~al., 2010]{le2010learning}
Le, Q.~V., Kamm, D., Kara, A.~F., and Ng, A.~Y. (2010).
\newblock Learning to grasp objects with multiple contact points.
\newblock In {\em ICRA}, pages 5062--5069.

\bibitem[Lenz et~al., 2015]{lenz2015deep}
Lenz, I., Lee, H., and Saxena, A. (2015).
\newblock Deep learning for detecting robotic grasps.
\newblock {\em IJRR}, 34(4-5):705--724.

\bibitem[Mairal et~al., 2009]{mairal2009online}
Mairal, J., Bach, F., Ponce, J., and Sapiro, G. (2009).
\newblock Online dictionary learning for sparse coding.
\newblock In {\em ICML}, pages 689--696.

\bibitem[Maitin-Shepard et~al., 2010]{maitin2010cloth}
Maitin-Shepard, J., Cusumano-Towner, M., Lei, J., and Abbeel, P. (2010).
\newblock Cloth grasp point detection based on multiple-view geometric cues
  with application to robotic towel folding.
\newblock In {\em ICRA}, pages 2308--2315.

\bibitem[Mar{\v{c}}elja, 1980]{marvcelja1980mathematical}
Mar{\v{c}}elja, S. (1980).
\newblock Mathematical description of the responses of simple cortical cells.
\newblock {\em JOSA}, 70(11):1297--1300.

\bibitem[Miller and Allen, 2004]{miller2004graspit}
Miller, A.~T. and Allen, P.~K. (2004).
\newblock Graspit!: A versatile simulator for robotic grasping.
\newblock {\em Robotics \& Automation Magazine}, 11(4):110--122.

\bibitem[Pati et~al., 1993]{pati1993orthogonal}
Pati, Y.~C., Rezaiifar, R., and Krishnaprasad, P. (1993).
\newblock Orthogonal matching pursuit: Recursive function approximation with
  applications to wavelet decomposition.
\newblock In {\em Signals, Systems and Computers. Twenty-Seventh Asilomar
  Conference on}, pages 40--44.

\bibitem[Pelossof et~al., 2004]{pelossof2004svm}
Pelossof, R., Miller, A., Allen, P., and Jebara, T. (2004).
\newblock An {SVM} learning approach to robotic grasping.
\newblock In {\em ICRA}, volume~4, pages 3512--3518.

\bibitem[Raina et~al., 2007]{raina2007self}
Raina, R., Battle, A., Lee, H., Packer, B., and Ng, A.~Y. (2007).
\newblock Self-taught learning: Transfer learning from unlabeled data.
\newblock In {\em ICML}, pages 759--766.

\bibitem[Redmon and Angelova, 2015]{redmonRealtimeCnnGrasp}
Redmon, J. and Angelova, A. (2015).
\newblock Real-time grasp detection using convolutional neural networks.
\newblock In {\em ICRA}, pages 1316--1322.

\bibitem[Rusu et~al., 2010]{rusu2010fast}
Rusu, R.~B., Bradski, G., Thibaux, R., and Hsu, J. (2010).
\newblock Fast 3{D} recognition and pose using the viewpoint feature histogram.
\newblock In {\em IROS}, pages 2155--2162.

\bibitem[Saxe et~al., 2011]{saxe2011random}
Saxe, A., Koh, P.~W., Chen, Z., Bhand, M., Suresh, B., and Ng, A.~Y. (2011).
\newblock On random weights and unsupervised feature learning.
\newblock In {\em ICML}, pages 1089--1096.

\bibitem[Saxena et~al., 2006]{saxena2006robotic}
Saxena, A., Driemeyer, J., Kearns, J., and Ng, A.~Y. (2006).
\newblock Robotic grasping of novel objects.
\newblock In {\em NIPS}, pages 1209--1216.

\bibitem[Saxena et~al., 2008]{saxena2008robotic}
Saxena, A., Driemeyer, J., and Ng, A.~Y. (2008).
\newblock Robotic grasping of novel objects using vision.
\newblock {\em IJRR}, 27(2):157--173.

\bibitem[Schmidt, 2005]{minFunc}
Schmidt, M. (2005).
\newblock minfunc: unconstrained differentiable multivariate optimization in
  matlab.

\bibitem[Socher et~al., 2012]{socher2012convolutional}
Socher, R., Huval, B., Bath, B., Manning, C.~D., and Ng, A.~Y. (2012).
\newblock Convolutional-recursive deep learning for 3{D} object classification.
\newblock In {\em NIPS}, pages 665--673.

\bibitem[Wright et~al., 2010]{wright2010sparse}
Wright, J., Ma, Y., Mairal, J., Sapiro, G., Huang, T.~S., and Yan, S. (2010).
\newblock Sparse representation for computer vision and pattern recognition.
\newblock {\em Proceedings of the IEEE}, 98(6):1031--1044.

\bibitem[Yang et~al., 2009]{yang2009linear}
Yang, J., Yu, K., Gong, Y., and Huang, T. (2009).
\newblock Linear spatial pyramid matching using sparse coding for image
  classification.
\newblock In {\em CVPR}, pages 1794--1801.

\bibitem[Zhang and Li, 2010]{zhang2010discriminative}
Zhang, Q. and Li, B. (2010).
\newblock Discriminative {K-SVD} for dictionary learning in face recognition.
\newblock In {\em CVPR}, pages 2691--2698.

\end{thebibliography}
}

\end{document}